\documentclass[conference]{IEEEtran}
\IEEEoverridecommandlockouts
\usepackage{cite}
\usepackage{amsmath,amssymb,amsfonts}
\usepackage{algorithmic}
\usepackage{graphicx}
\usepackage{textcomp}
\usepackage{xcolor}

\usepackage{algorithm}
\usepackage{algorithmic}
\usepackage{pifont}

\usepackage{tabularray}
\usepackage{multirow}
\usepackage{booktabs}
\usepackage{url}

\def\BibTeX{{\rm B\kern-.05em{\sc i\kern-.025em b}\kern-.08em
    T\kern-.1667em\lower.7ex\hbox{E}\kern-.125emX}}
    
\begin{document} 


\title{Slicing Input Features to Accelerate Deep Learning: A Case Study with Graph Neural Networks}

\DeclareRobustCommand*{\IEEEauthorrefmark}[1]{%
  \raisebox{0pt}[0pt][0pt]{\textsuperscript{\footnotesize #1}}%
}
\author{
    \IEEEauthorblockN{ 
        Zhengjia Xu\IEEEauthorrefmark{1}, 
        Dingyang Lyu\IEEEauthorrefmark{2},
        Jinghui Zhang\IEEEauthorrefmark{2}\textsuperscript{\textasteriskcentered}
    }
    \IEEEauthorblockA{
        \IEEEauthorrefmark{1}College of Software Engineering, Southeast University, China\\
        \IEEEauthorrefmark{2}School of Computer Science and Engineering, Southeast University, China\\
        Email: 
            \{xzhengjia, dylv, jhzhang\}@seu.edu.cn
    }
    \thanks{* Corresponding Author}
}

\maketitle
\thispagestyle{plain} 

\begin{abstract}
As graphs grow larger, full-batch GNN training becomes hard for single GPU memory. Therefore, to enhance the scalability of GNN training, some studies have proposed sampling-based mini-batch training and distributed graph learning. However, these methods still have drawbacks, such as performance degradation and heavy communication. This paper introduces SliceGCN, a feature-sliced distributed large-scale graph learning method. SliceGCN slices the node features, with each computing device, i.e., GPU, handling partial features. After each GPU processes its share, partial representations are obtained and concatenated to form complete representations, enabling a single GPU's memory to handle the entire graph structure. This aims to avoid the accuracy loss typically associated with mini-batch training (due to incomplete graph structures) and to reduce inter-GPU communication during message passing (the forward propagation process of GNNs). To study and mitigate potential accuracy reductions due to slicing features, this paper proposes feature fusion and slice encoding. Experiments were conducted on six node classification datasets, yielding some interesting analytical results. These results indicate that while SliceGCN does not enhance efficiency on smaller datasets, it does improve efficiency on larger datasets. Additionally, we found that SliceGCN and its variants have better convergence, feature fusion and slice encoding can make training more stable, reduce accuracy fluctuations, and this study also discovered that the design of SliceGCN has a potentially parameter-efficient nature.
\end{abstract}

\begin{IEEEkeywords}
Deep Learning, Distributed Learning, GNN
\end{IEEEkeywords}

\section{Introduction}

With the rapid development of deep learning and the accumulation of large amounts of graph data, the application of deep learning on graphs has become a focal point in research, with promising applications in fields like transportation networks \cite{febrinanto2023graph}, molecular analysis \cite{rao2014protein}, and financial transactions \cite{huang2022dgraph}. Graph Neural Networks (GNNs) \cite{gnn4700287} learn node representations by aggregating information from neighboring nodes into target nodes through message passing iteratively, which has become the de facto standard method for graph learning.

\begin{figure}[!t]
    \centering
    \includegraphics[width=1\linewidth]{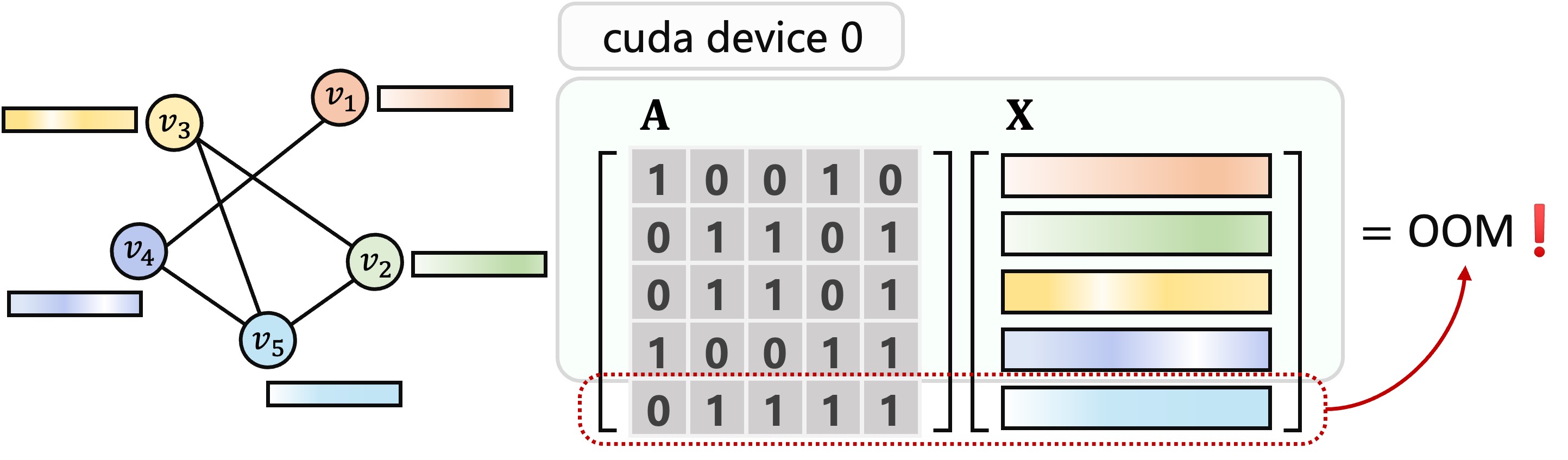}
    \caption{A simple example of the memory shortage issue in GNN full-batch learning.}
    \label{fig:1_fullbatch}
\end{figure}

In GNNs forwarding, as the number of layers (i.e., the number of message-passing iterations) increases, a node requires information from a higher-order neighborhood. Therefore, the basic training method for GNNs involves inputting all node features in the graph at once (full-batch), meaning that when a node require farther neighbors, it can access since the required high-order nerighbors must be in the full batch. This is simple, but not suitable for large graphs. As shown in Figure \ref{fig:1_fullbatch}, placing large-scale graph data into the memory of a single acceleration device (such as a GPU) for full-batch learning will cause Out of Memory (OOM). To cope with large graphs, researchers employ methods like graph sampling and distributed training.

\begin{figure}[!t]
    \centering
    \includegraphics[width=0.79\linewidth]{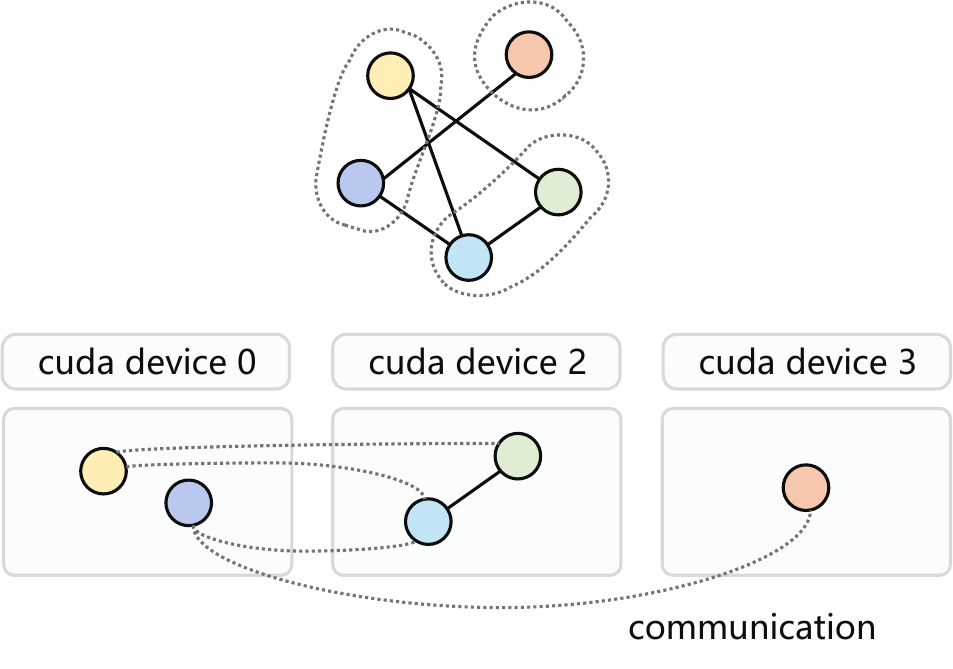}
    \caption{Distributed graph learning with subgraph partitioning.}
    \label{fig:1_distgnn}
\end{figure}

Through graph sampling \cite{hamilton2017inductive}, mini-batch training can be performed, allowing for smaller memory usage, but after just a few layers of message passing, the number of required nodes grows exponentially, that is, nodes in a mini-batch will eventually depend on a large number of neighboring nodes or even the entire graph. Therefore, limiting the number of sampled nodes may result in some nodes in a mini-batch not being fully trained, leading to feature approximation errors and ultimately causing a decline in learning quality \cite{wang2023mgg}\cite{wan2022pipegcn}. Distributed computing can train on large graphs using a full-batch approach.
Common distributed learning methods include data parallelism and model parallelism. As previously analyzed, mini-batch learning compromises accuracy; therefore, data parallelism, which relies on mini-batches, is unsuitable for graph learning. On the other hand, model parallelism and related pipeline training methods, such as GPipe \cite{huang2019gpipe} and PipeDream \cite{narayanan2019pipedream}, are appropriate for large models with many layers, whose parameters cannot be accommodated by a single GPU and require distributed placement across multiple GPUs. However, unlike Convolutional Neural Networks (CNNs) and Transformers, which significantly benefit from increasing the number of layers to tens or more, GNNs are typically shallower. Deeper layers in GNNs can lead to issues like oversmoothing \cite{10.1145/3394486.3403076} and over-squashing \cite{topping2022understanding}, suggesting that GNN training bottlenecks are not related to model parameter size. Thus, model parallelism and pipeline parallelism might also not be suitable for GNN training.
Therefore, a common strategy in distributed graph learning is to cut a graph into subgraphs and distribute them across computing devices \cite{wan2022pipegcn}. During training, transferring graph nodes between computing devices (such as GPUs) ensures the integrity of the graph, thereby achieving full-batch accuracy. However, as shown in Figure \ref{fig:1_distgnn}, the potential heavy communication overhead may be an efficiency issue.

In this paper, we propose SliceGCN which requires less device communication, a distributed graph learning method based slicing node features. Input features are sliced along the feature dimension to obtain feature slices, each processed by a device, allowing each device to utilize the complete graph structure and feature slices for training. During forward propagation, there is no need for exchanging training information between devices. At the beginning and end of SliceGCN, feature fusion and slice encoding are designed to ensure model accuracy, since incomplete features in forward propagation may impair performance. Communication needs are only present at the beginning (transfering the sliced graphs to each device) and end of the process (concatenating the representations output by the distributed devices). Experimental results show that SliceGCN maintains or even improves model accuracy, while enhancing training efficiency on large-scale graph data.
Our contributions are as follows:

\begin{itemize}
    \item A feature-sliced distributed graph learning method, SliceGCN, is proposed. We design a simple slicing strategy and parameter initialization for SliceGCN. By slicing node features and distributing them across devices, larger graphs can be processed using full-batch manner on a single device.

    \item To ensure that the training performance is not impaired by the incompleteness of sliced features, feature fusion and slice encoding are employed.

    \item Experiments on six public node classification datasets demonstrate the effectiveness of SliceGCN. Additionally, ablation studies of slice encoding and feature fusion are conducted.
    
\end{itemize}

\section{Related Work}

As the scale of deep learning models and training data grows, distributed learning methods have been explored to improve the scalability and training efficiency. These methods can be categorized into data parallelism, model parallelism, and pipeline parallelism \cite{10.1145/3377454}. Data parallelism is the simplest and most widely used. It copies the model parameters to multiple devices (e.g., GPUs) and synchronizes the gradients of each device to update the parameters, enabling training on more data simultaneously. Model parallelism partitions the model into multiple parts, with each device responsible for computing the output of a portion of the layers and transmitting it to the next device, eventually backpropagating from the last device to the input layer. Pipeline parallelism further improves the efficiency of model parallelism by introducing a pipeline-like training method \cite{huang2019gpipe, narayanan2019pipedream}. For example, in GPipe \cite{huang2019gpipe}, the model is divided into multiple parts distributed across devices (model parallelism), and the mini-batch is further divided into smaller micro-batches. After the micro-batch is processed by the first device, the output is passed to the next device, which waits for the subsequent device to complete the forward propagation before backpropagating. In naive model parallelism, the first device is idle at this time. Therefore, GPipe uses the idle device to process the next micro-batch, forming a pipeline parallelism that improves the utilization of computing resources. PipeDream \cite{narayanan2019pipedream} further improves the utilization of computing resources in pipeline training by introducing asynchronous computing.
These classic distributed learning methods have been widely used and have greatly promoted the advancement of deep learning, but they are not suitable for GNNs. 
There are distributed GNN training methods based on sampling, such as PaGraph \cite{pagraph}, which is sampling-based mini-batch graph learning. PaGraph points out the impact of data movement between CPUs and GPUs on efficiency and improves efficiency by caching the features of the most frequently accessed nodes. However, some studies have shown that sampling can compromise accuracy \cite{wang2023mgg}. Data parallelism requires data to be divided into mini-batches, but graph data is difficult to partition without performance loss due to the dependencies between nodes. Unlike general training samples, the dependencies between samples (nodes) in graph data can provide rich topological information beyond features, and will be lost during graph sampling. Therefore, GNNs achieve better results with full-batch learning.
Model parallelism is suitable for models with a large number of parameters, but the parameter scale of GNNs is relatively small. When the number of layers is too deep, problems like gradient vanishing and neighborhood explosion may occur. Although there are works exploring deep GNNs \cite{deepgcn,li2020deepergcn}, mainstream GNNs still employ shallow networks. Pipeline parallelism is also suitable for models with a large number of parameters and input data that can be divided into mini-batches, making it unsuitable for GNNs.

When the memory of a single device (e.g., GPU memory) is insufficient to process an entire graph, a typical distributed GNN training method divides the graph into small subgraphs distributed across devices, and during training, communicates between devices to transfer boundary nodes, i.e., neighboring nodes at the boundaries of subgraphs, to achieve full-batch training and maintain the dependencies between nodes. These methods explore optimization from the perspective of node data transmission.
For example, NeuGraph \cite{234948} uses a 2D partitioning algorithm to divide the original graph into multiple subgraph blocks, with each node's neighbors divided into multiple groups and executed sequentially to obtain the final result. NeuGraph constructs a flow scheduler to overlap data transmission and computation to improve efficiency. ROC \cite{jia2020improving} designs a learnable online graph partitioning method to predict the execution time of the GNN layer after partitioning and jointly trains with the GNN to minimize the execution time of the GNN layer after partitioning, achieve load balancing, and improve efficiency. ROC is suitable for both single-machine multi-GPU and GPU server clusters.
DGCL \cite{10.1145/3447786.3456233} is a GNN training framework designed for GPU server clusters, which particularly considers the physical communication topology to determine the optimal communication strategy. PipeGCN \cite{wan2022pipegcn} is a full-batch distributed training system developed based on DistDGL \cite{zheng2020distdgl}. It constructs a ``pipeline'' system that overlaps node information transmission and GNN layer computation (not the pipeline system based on model parallelism introduced earlier). For neighboring nodes on other computing devices, PipeGCN uses an asynchronous delayed update method, i.e., using the information of neighboring nodes transmitted in the previous round during computation, to overlap computation and communication. However, this introduces outdated information, so PipeGCN introduces the design of outdated feature gradient smoothing to improve this issue. Dorylus \cite{273743} designs a distributed GNN system based on CPU servers, dividing the graph into different partitions using an cutting-edge algorithm, and sending and receiving partition boundary data between servers during computation, demonstrating performance comparable to GPU servers and higher cost-effectiveness. Unlike the above methods, P3 \cite{273707p3} divides features and uses data parallelism to process model parameter synchronization. The above methods require a large amount of communication during computation to ensure accuracy. Asynchronous communication methods can improve efficiency, but also introduce a certain degree of accuracy loss.

\section{Preliminaries}

\subsection{Attributed Graph}
$\mathcal{G}=\{\mathcal{V}, \mathbf{X}, \mathcal{E}, \mathcal{Y}, \mathcal{C}\}$ is a graph, where $\mathcal{V}= \{v_1, v_2, \ldots, v_{n}\}$ is the node set, $n=\left|\mathcal{V}\right|$ is the number of nodes, and $\mathcal{C}$ is the node category set.
For each node $v_{i} \in \mathcal{V}$, there is a neighbor set $\mathcal{N}(v_i)$. $\mathcal{E}$ is the edge set, $\mathcal{E}=\{(v_{i},v_{j}) | v_{i},v_{j}\in\mathcal{V}\}$. The edge set can also be represented by an adjacency matrix $\mathbf{A} \in \{0, 1\}^{n \times n}$, where $\mathbf{A}_{ij} = 1$ indicates that there is an edge between nodes $v_{i}$ and $v_{j}$, otherwise $\mathbf{A}_{ij} = 0$.
For each node $v_{i}\in\mathcal{V}$, there is a label $y_{i} \in {\mathcal{Y}}$, where $\mathcal{Y}= \{y_1, y_2, \ldots, y_{n}\}$ is the node label set, satisfying $\mathcal{Y} \subseteq \mathbb{N}$ and for $\forall y_{i}$, $y_{i} \in [0, |\mathcal{C}|)$, corresponding to the one-hot encoding representation of node labels $\mathbf{Y}= \{\mathbf{y}_1, \mathbf{y}_2, \ldots, \mathbf{y}_{n}\}$, where $\mathbf{Y}\in\mathbb{R}^{\left|\mathcal{V}\right| \times |\mathcal{C}|}$. $\mathbf{X}=\{\mathbf{x}_{1},\mathbf{x}_{2},...,\mathbf{x}_{n}\}\in\mathbb{R}^{n \times d'}$ is the corresponding node feature, i.e., node attribute, where $d'$ is the feature dimension (not the node representation dimension $d$ in the following text).

\subsection{Graph Convolutional Network}
The goal of GNN is to learn node representations $\mathbf{H} \in \mathbb{R}^{n \times d}= \{\mathbf{h}_{1}, \mathbf{h}_{2}, \ldots, \mathbf{h}_{n}\}$ for downstream tasks, given a graph $\mathcal{G}=\{\mathcal{V}, \mathbf{X}, \mathcal{E}, \mathcal{Y}, \mathcal{C}\}$, where $d$ is the dimension of node representations. This paper takes the Graph Convolutional Network (GCN) as an example. The graph convolutional layer can be defined by the following:
\begin{equation}
\label{eq:gcn}
    \mathbf{h}_{i}^{(l+1)} = \mathrm{ReLU} \left(\mathbf{b}^{(l)} + \sum_{j \in \mathcal{N}(v_i)} \frac{1}{c_{ij}} \mathbf{h}_{j}^{(l)} \mathbf{W_\mathrm{agg}}^{(l)}\right)
\end{equation}
where $\mathcal{N}(v_i)$ is the neighbor set of node $v_i$, $\mathrm{ReLU} \left( \cdot \right) $ is the activation function. $\mathbf{b}^{(l)}$ is the bias vector, $\mathbf{W_\mathrm{agg}}^{(l)}$ is the learnable parameter of the $l$-th layer. $c_{ij}$ is the product of the square root of the degrees of nodes, i.e., $c_{ij} = \sqrt{|\mathcal{N}(v_i)|} \sqrt{|\mathcal{N}(v_j)|}$, used to normalize the information transmission of neighbors.

\subsection{Distributed GCN Training Task}

Given a graph $\mathcal{G}=\{\mathcal{V}, \mathbf{X}, \mathcal{E}, \mathcal{Y}, \mathcal{C}\}$ and $p$ computing devices (GPUs), the goal is to distribute the training of GCN to train larger-scale graph data that cannot be handled by a single computing device (e.g., a single GPU due to memory limitations) in full-batch manner. This task requires partitioning the graph data and distributing it across computing devices while ensuring model convergence.

\begin{figure}[!tbp]
    \centering
    \includegraphics[width=1\linewidth]{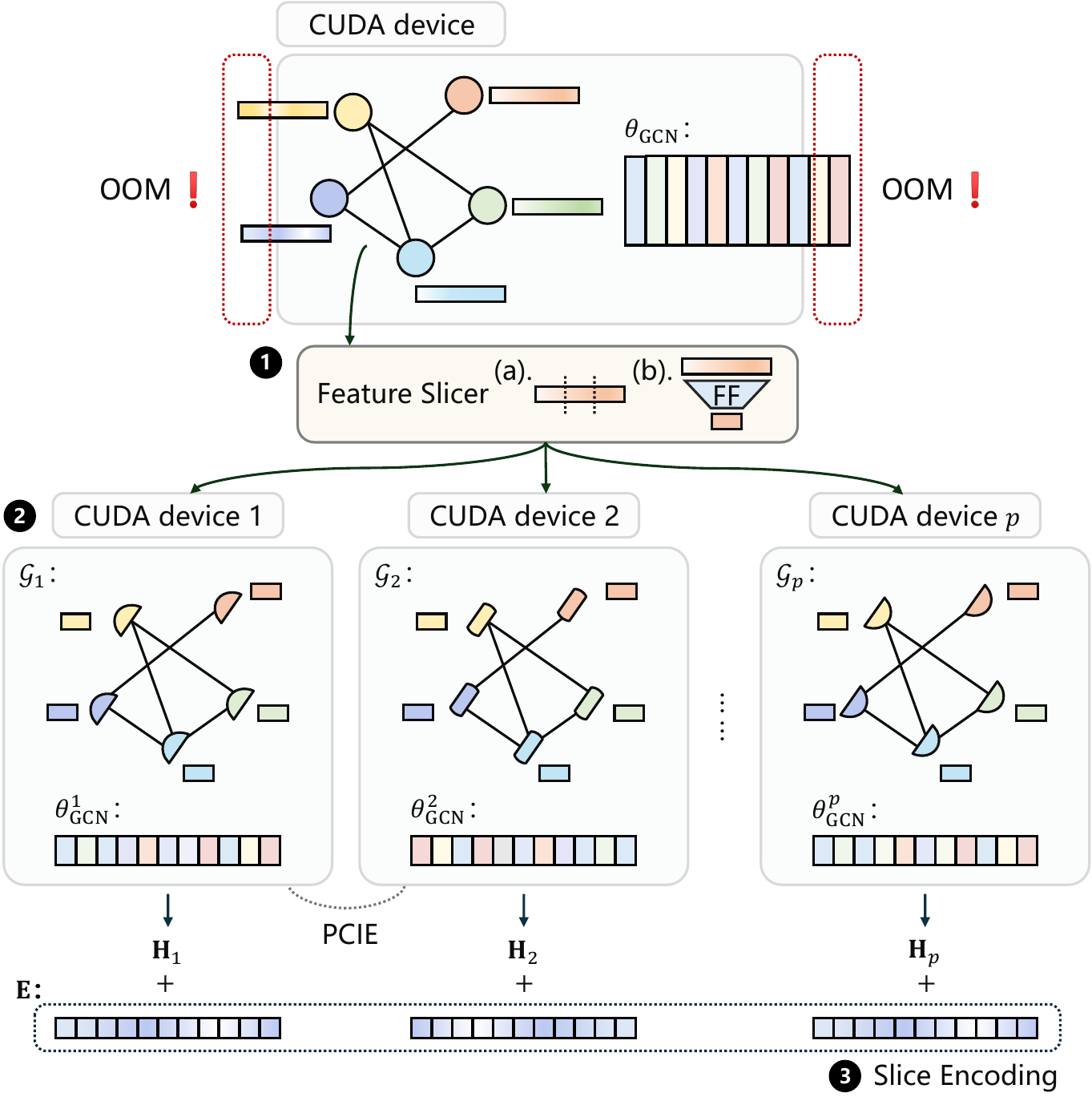}
    \caption{The framework of SliceGCN.}
    \label{fig:slice-gcn}
\end{figure}

\section{SliceGCN}

Figure \ref{fig:slice-gcn} shows the framework of SliceGCN. The input features are sliced into $p$ parts, corresponding to $p$ GPUs, each of which loads the complete graph structure and a small slice of features for all nodes, reducing memory consumption. On the $i$-th GPU, there is a set of GCN parameters, ${\theta}_{\mathrm{GCN}}^i$, with smaller scale due to the smaller node features. In each iteration, each GPU performs forward propagation on its feature slice, with communication between GPUs only occurring before the input of graph data and after the output of GCN, 
As shown in Figure \ref{fig:slice-gcn} \ding{183}, unlike Figure \ref{fig:1_distgnn}, in SliceGCN, the parallel computing process is completed independently without communication.
At the end of SliceGCN, as shown in Figure \ref{fig:slice-gcn} \ding{184}, the GCN model on each GPU will transfer its output representation $\mathbf{H}_{i}$ to the master node, which will concatenate the output representations of each GPU to obtain the complete node representation $\mathbf{H}$, then input it to the classifier for downstream tasks. Considering the relative independence of each set of GCN parameters $\theta_{i}$, we design slice encoding before aggregation to adjust the output representations of each GCN to improve the quality of the concatenated representations. The following sections will provide a detailed introduction to SliceGCN.

\subsection{Feature Slicing and Model Initialization}

At the beginning of SliceGCN, as shown in Figure \ref{fig:slice-gcn} \ding{182}, two feature slicing methods are designed: (a) direct slicing and (b) feature fusion. The purpose of both methods is to process the input features $\mathbf{X} \in \mathbb{R}^{n \times d'}$ to obtain feature slices $\mathbf{X}^i \in \mathbb{R}^{n \times \frac{d'}{p}}$ with smaller storage space, where $n$ is the number of nodes, $d'$ is the feature dimension, and $p$ is the number of computing devices.
In fact, simply put, feature fusion reduces the dimensionality of raw features to obtain slices, whereas direct slicing cuts directly.
In (a) direct slicing, the input features $\mathbf{X}$ are simply divided into $p$ parts. The calculation of $\mathbf{X}^i$ ($i \in [1, p]$) is as follows:
\begin{equation}
    \mathbf{X}^i = \mathbf{X}[:, start : {end}]
    \label{eq:x_slice}
\end{equation}
where the calculation of $start$ and $end$ is as follows:
\begin{equation}
    \begin{aligned}
     { start } & = \begin{cases}(i-1) \cdot\left\lceil\frac{d'}{p}\right\rceil+1,& \text { if } i \cdot\left\lceil\frac{d'}{p}\right\rceil<d' \\
    d-\left\lceil\frac{d'}{p}\right\rceil+1,& \text { otherwise }\end{cases} \\
     { end } & = \begin{cases}i \cdot\left\lceil\frac{d'}{p}\right\rceil,& \text { if } i \cdot\left\lceil\frac{d'}{p}\right\rceil<d' \\
    d',& \text { otherwise }\end{cases}
    \end{aligned}
\end{equation}

Therefore, SliceGCN has APIs interfaces \path{slice_strategy_generator()} and \path{slice_feature()}, designed as shown in Algorithm \ref{alg:slice_strategy} and Algorithm \ref{alg:feature_slice}. Here, \path{slice_size_scale} is the slice size scaling factor, which can be used to adjust the size of the feature slice on each GPU to adapt to different computing resources. The default setting is 1.0.

\begin{algorithm}[!t]
    \caption{Slice Strategy Generation}
    \begin{algorithmic}[1]
    \label{alg:slice_strategy}
        \REQUIRE Feature dimension \texttt{in\_d}, number of computing devices \texttt{p}, slice size scaling factor \texttt{slice\_size\_scale}
        \ENSURE Slice strategy and slice size
        \STATE Initialize slice list \texttt{in\_sizes} $\leftarrow$ \texttt{List()}
        \STATE \texttt{slice\_size} $\leftarrow$ $\lceil$ \texttt{in\_d} / \texttt{p} $\rceil$
        \FOR{ \texttt{i} $= 0, 1, \ldots, \texttt{p}-1$ }
            \STATE \texttt{slice\_start} $\leftarrow$ \texttt{i} $\times$ \texttt{slice\_size} 
            // Calculate the start position of the slice.
            \STATE \texttt{slice\_end} $\leftarrow$ \path{slice_start} + \path{int(slice_size} $\times$ \path{slice_size_scale)} \\
            // Calculate the end position of the slice.
            \IF{\texttt{slice\_end} $>$ \texttt{in\_d}} 
            \STATE // Handle the last slice.
                \STATE \texttt{slice\_start} $\leftarrow$ \texttt{slice\_start} - (\texttt{slice\_end} - \texttt{in\_d}) // Adjust the start position of the slice.
                \STATE \texttt{slice\_end} $\leftarrow$ \texttt{in\_d} // Set the end position of the slice to the maximum value of the feature dimension.
            \ENDIF
            \STATE \texttt{in\_sizes.append((slice\_start, slice\_end))} // Add the slice strategy.
        \ENDFOR
        \STATE \textbf{return} \texttt{in\_sizes}, \texttt{slice\_size}
    \end{algorithmic}
\end{algorithm}

\begin{algorithm}[!thbp]
    \caption{Feature Slicing}
    \begin{algorithmic}[1]
    \label{alg:feature_slice}
        \REQUIRE Input feature \texttt{x}, slice strategy \texttt{in\_sizes}
        \ENSURE List of sliced features
        \STATE Initialize sliced feature list \texttt{sliced\_features} $\leftarrow$ \texttt{List()}
        \FOR{\texttt{i}, (\path{slice_start}, \path{slice_end}) \textbf{in} \path{enumerate(in_sizes)}}
            \STATE \path{sliced_feature} $\leftarrow$ \path{x[:, slice_start:slice_end].to(devices[i])}
            \STATE \texttt{sliced\_features.append(sliced\_feature)} // Add the sliced feature to the list.
        \ENDFOR
        \STATE \textbf{return} \texttt{sliced\_features}
    \end{algorithmic}   
\end{algorithm}

In (b) feature fusion, the calculation of $\mathbf{X}^i$ is as follows:
\begin{equation}
    \mathbf{X}^i = \mathbf{X} \mathbf{W}_\mathrm{FF} 
    \label{eq:x_slice_ff}
\end{equation}
where $\mathbf{W}_\mathrm{FF} \in \mathbb{R}^{d \times \left(\left\lceil{\frac{d'}{p}}\right\rceil + 1 \right)}$ is the learnable parameter matrix for feature fusion.

After processing the input features, the SliceGCN models are initialized: for the $i$-th GPU, the learnable aggregation weight $\mathbf{W}_\mathrm{agg}^{i,(l)} \in \mathbb{R}^{d \times d}$ and the self-update weight $\mathbf{W}_\mathrm{self}^{i,(l)} \in \mathbb{R}^{d \times d}$ ($\mathbf{W}_\mathrm{self}^{1,(l)} \in \mathbb{R}^{ \left(\left\lceil{\frac{d'}{p}}\right\rceil + 1 \right) \times d}$) are initialized for the $l$-th layer of the GCN model on the $i$-th GPU. For the $l$-th layer of node $v \in \mathcal{V}$ on the $i$-th GPU, the calculation is as follows:
\begin{equation}
    \label{eq:45}
        \mathbf{h}_{v}^{i,(0)} = \mathbf{x}_{v}^i
\end{equation}
\begin{equation}
    \label{eq:46}
    \begin{aligned}
        \mathbf{h}_{v}^{i,(l)} = &\mathrm{ReLU}\left(\mathbf{b}^{i,(l)} + \sum_{u \in \mathcal{N}(v)} \frac{1}{c_{uv}} \mathbf{h}_{u}^{i,(l-1)} \mathbf{W}_\mathrm{agg}^{i,(l)} \right)
        + \\&\mathbf{h}_{v}^{i,(l-1)} \mathbf{W}_{\mathrm{self}}^{i,(l)}
    \end{aligned}
\end{equation}

As shown in Figure \ref{fig:slice-gcn} \ding{183}, after the above calculations are completed in parallel on $p$ computing devices, $p$ sets of node representations $\mathbf{H}^{i,(L)}$ are obtained, where $L$ is the number of layers in the GCN.



\subsection{Slice Encoding and Model Ouput}

Considering that the $p$ GCN models, without communication or interaction in the middle, may learn different representation patterns, we design slice encoding, as shown in Figure \ref{fig:slice-gcn} \ding{184}, to adjust the output representations of each GCN. The process of adjusting the representation output of the $i$-th computing device is as follows:
\begin{equation}
    \label{eq:47}
        \mathbf{H}^i = \mathbf{H}^{i,(L)} + \mathbf{E}_{1, :}
\end{equation}
where $\mathbf{E}_{1, :} \in \mathbb{R}^{p \times d}$ is the parameter matrix of slice encoding. Finally, the master node concatenates the output representations $\mathbf{H}^i$ of each GCN and inputs them into the MLP classifier responsible for downstream tasks to complete the forward propagation process and backpropagation training:
\begin{equation}
    \label{eq:48}
        \mathbf{H} = \text{CONCAT}(\mathbf{H}^1, \mathbf{H}^2, \ldots, \mathbf{H}^p)
\end{equation}
\begin{equation}
    \label{eq:49}
        \hat{\mathbf{Y}}_{L} = \mathrm{MLP}_{\text{CLS}}(\mathbf{H}_{L})
\end{equation}
\begin{equation}
    \label{eq:50}
        \mathcal{L} = \text{CE}(\hat{\mathbf{Y}}_{L}, \mathbf{Y}_L)
\end{equation}

The training process of SliceGCN is shown in Algorithm \ref{alg:slice-gcn-train}.
 
\begin{algorithm}[!ht]
    \caption{Training Process of Slice-GCN}
    \begin{algorithmic}[1]
    \label{alg:slice-gcn-train}
        \REQUIRE Graph data $\mathcal{G}=\{\mathcal{V}, \mathbf{X}, \mathcal{E}, \mathcal{Y}, \mathcal{C}\}$, training set $\mathcal{V}_L$, training set labels $\mathbf{Y}_{\mathrm{L}}$, SliceGCN layers $L$, number of epochs $E$, number of computing devices $p$, slice size scaling factor \path{slice_size_scale} = 1.0
        \ENSURE Trained node representation $\mathbf{H}$
        \STATE \path{in_sizes}, \path{slice_size} $\leftarrow$ \path{slice_strategy_generator} $(d', p,$ \path{slice_size_scale}$=1.0)$
        \STATE Initialize SliceGCN parameters on $p$ computing devices
        \STATE Initialize Slice Encoding parameters
        \STATE Initialize $\mathrm{MLP}_{\text{CLS}}$ parameters
        \IF {\path{use_ff}}
            \STATE $\mathrm{MLP}_{\mathrm{FF}}$ $\leftarrow$ \path{init_mlp_ff}$($\path{in_size}$=d', $\path{h_size}$=d', $\path{out_size}$=$\path{slice_size}$, $\path{layers}$=2)$
        \ELSE
            \STATE \path{x_list} $\leftarrow$ \path{slice_feature}$(\mathbf{X}, $\path{in_sizes}$)$
        \ENDIF
        \FOR {$epoch = 1, 2, \ldots, E$}
            \IF {\path{use_ff}}
                \STATE \path{x_list} $\leftarrow$ $\mathrm{MLP}_{\mathrm{FF}}$$(\mathbf{X})$
            \ENDIF
            \FOR {$i = 1, 2, \ldots, p$} 
            \STATE // Start parallel processing
                \FOR {$v \in \mathcal{V}$}
                    \STATE $\mathbf{h}_{v}^{i,(0)} \leftarrow \mathbf{x}_{v}^i$
                \ENDFOR
                \FOR {$l = 1, 2, \ldots, L$}
                    \STATE $\mathbf{h}_v^{i,(l)} \leftarrow$ Eq. \ref{eq:46}
                \ENDFOR
                \STATE $\mathbf{H}^i \leftarrow$ Eq. \ref{eq:47}
            \ENDFOR
            \STATE $\mathbf{H} \leftarrow \text{CONCAT}(\mathbf{H}^1, \mathbf{H}^2, \ldots, \mathbf{H}^p)$
            \STATE $\hat{\mathbf{Y}}_{L} \leftarrow$ $\mathrm{MLP}_{\text{CLS}}$$(\mathbf{H})$
            \STATE $\mathcal{L} \leftarrow \text{CE}(\hat{\mathbf{Y}}_{L}, \mathbf{Y})$
            \STATE Backpropagation to update parameters
        \ENDFOR
        \STATE \textbf{return} Trained node representation $\mathbf{H}$
    \end{algorithmic}
\end{algorithm}

\section{Experiments}

\subsection{Datasets}

The datasets used to test the effectiveness of SliceGCN are shown in Table \ref{tab:datasets-slice}. Roman-Empire, Amazon-Ratings, Minesweeper, Tolokers, and Questions are heterophilous graph datasets proposed by Platonov et al. \cite{platonov2023a_new_dataset}. Roman-Empire is a network of Roman-related entries on Wikipedia, Amazon-Ratings is a network of Amazon product ratings, Minesweeper is a network constructed from Minesweeper game data, Tolokers is a network of crowdsourcing platforms, and Questions is a user network from the Yandex Q question-and-answer website. 
OGBN-Arxiv is proposed by Open Graph Benchmark (OGB) \cite{ogbn}, a larger-scale academic paper citation network dataset, where each node is a paper, the edges represent citation relationships between papers, and each node has a 128-dimensional feature vector generated from the paper title and abstract.

\begin{table}[!ht]
    \caption{Statistics of datasets used in this paper.}
    \centering
    \begin{tabular}{c|ccccc} 
    \toprule
    Dataset & \#Nodes & \#Edges & \#Features & \#Classes \\
    \midrule
    Roman-Empire & 22662 & 32927 & 300 & 18 \\
    Amazon-Ratings & 24492 & 93050 & 300 & 5 \\
    Minesweeper & 10000 & 39402 & 7 & 2 \\
    Tolokers & 11758 & 519000 & 10 & 2 \\
    Questions & 48921 & 153540 & 301 & 2 \\
    \midrule
    OGBN-Arxiv & 169343 & 1166243 & 128 & 40 \\
    \bottomrule
    \end{tabular}
    \label{tab:datasets-slice}
\end{table}

\subsection{Compared Methods}
 
\begin{itemize} 
    
    \item {\textbf{Single-GPU GCN:}} Train the GCN model on a single GPU as the baseline method for SliceGCN.
    \item {\textbf{SliceGCN:}} The basic form of the distributed GCN training method proposed in this paper, which slices the input features and distributes them to various computing devices for parallel computation without feature fusion or slice encoding.
    \item {\textbf{SliceGCN-SE:}} Based on SliceGCN, the slice encoding module is added to adjust the output representation patterns of each computing device.
    \item {\textbf{SliceGCN-FF:}} Based on SliceGCN, feature fusion is used to process the input features, allowing each computing device to see relatively complete node feature information.
    \item {\textbf{SliceGCN-FFSE:}} Based on SliceGCN-FF, the slice encoding module is further added to adjust the output representation patterns of each computing device.
\end{itemize}

\subsection{Experimental Settings}

The proposed SliceGCN and the GCN baseline are implemented using Python 3.11.8, PyTorch 2.2.1, and DGL 2.1.0. All experiments were run on a server with 3 $\times$ NVIDIA GeForce RTX 3090 GPUs, 2 $\times$ 12-core Intel(R) Xeon(R) Silver 4214 CPUs @ 2.20GHz, 192 GB of RAM, Ubuntu 20.04.6 LTS, and CUDA 11.8.

All experiments employ the Adam optimizer and cosine annealing learning rate scheduler. The specific experimental configurations and hyperparameters for each dataset are detailed in Table \ref{tab:hyperparameters-slice-gcn}.

\begin{figure}[!tbp]
    \centering
    \includegraphics[width=1\linewidth]{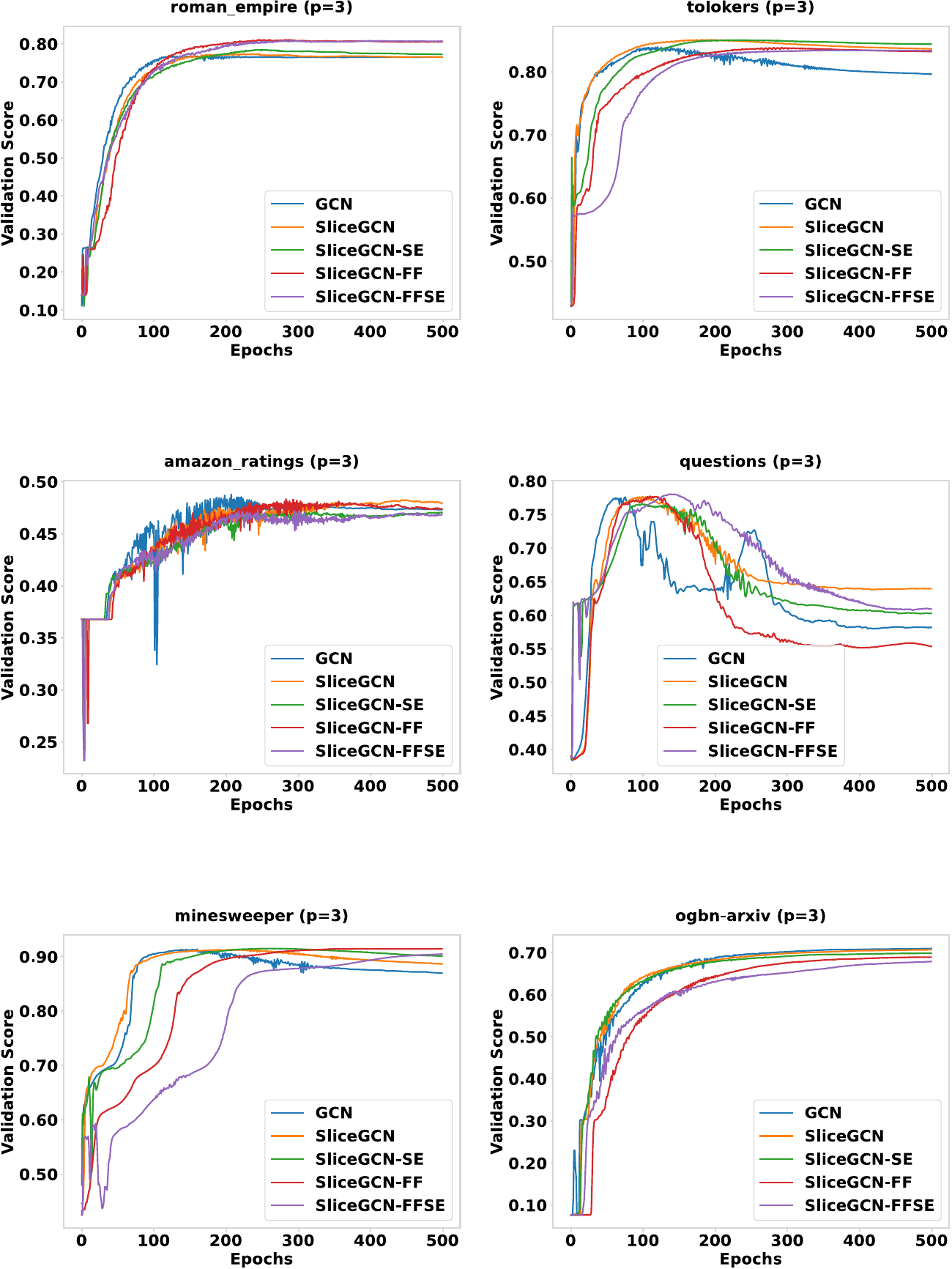} 
    \caption{Validation accuracy of GCN, SliceGCN, and its variants on different datasets with $p=3$.}
    \label{fig:SliceGCN-p3} 
\end{figure}

\section{Analysis}

 The number of parameters for GCN on a single GPU and for SliceGCN and its variants with different numbers of GPUs $p$ are shown in Table \ref{tab:params-stat-slice-gcn}. It can be seen that as the number of GPUs increases, the number of parameters decreases. This is because SliceGCN proportionally scales down each layer of the GCN model on each GPU according to $p$.

As shown in Table \ref{tab:res-slice-gcn}, the experimental results for single-GPU GCN and the proposed method SliceGNN on each dataset are presented. The accuracy is the highest test accuracy at the highest validation set accuracy, and the throughput is the number of epochs that can be processed per second. For SliceGCN and its variants, we tested two GPU quantities, $p=2$ and $p=3$, to observe the performance under different computational resources. For binary classification tasks, the AUC-ROC value is reported; for multi-class tasks, the accuracy is reported.

\begin{figure}[!tbp]
    \centering
    \includegraphics[width=1\linewidth]{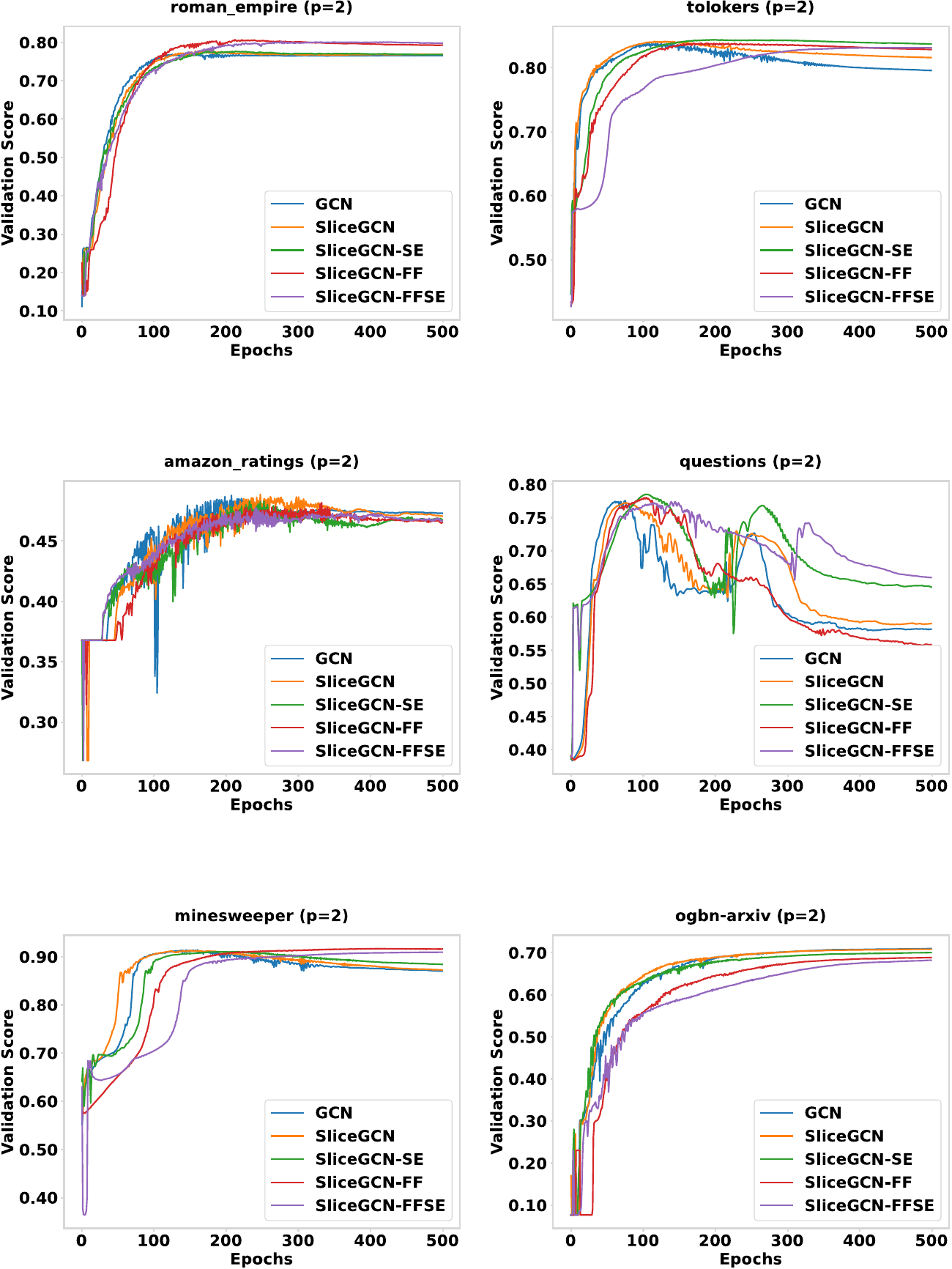} 
    \caption{Validation accuracy of GCN, SliceGCN, and its variants on different datasets with $p=2$.}
    \label{fig:SliceGCN-p2} 
\end{figure}

\begin{table}[!t]
    \caption{Hyperparameters settings for SliceGCN on different datasets.}
    \centering
    \begin{tabular}{l|ccccc} 
    \toprule
    Dataset  & \#Epochs & \#Hidden & $L$ & LR & Dropout  \\
    \midrule 
    Roman-Empire  & 500  & 256 & 3    & 0.001 & 0.5       \\
    Amazon-Ratings     & 500  & 128 & 4    & 0.001 & 0.5       \\
    Minesweeper   & 500  & 128 & 7    & 0.001 & 0.5     \\
    Tolokers   & 500  & 128 & 4    & 0.001 & 0.5      \\
    Questions  & 500  & 512 & 6    & 0.001 & 0.5     \\ \midrule
    OGBN-Arxiv  & 500  & 512 & 4   & 0.001 & 0.0     \\
    \bottomrule
    \end{tabular}
    \label{tab:hyperparameters-slice-gcn}
\end{table}

 \begin{table}[!tbp]
    \caption{Statistics of parameter counts for single-GPU GCN, SliceGCN, and SliceGCN variants on each dataset.}
    \centering
    \resizebox{\linewidth}{!}
    {
    \begin{tabular}{c|l|cccccc} 
    \toprule
    \multicolumn{2}{c|}{Method/Dataset}     & Rom.    & Amz.    & Mine. & Tol.    & Ques. & OGB.     \\ 
    \midrule
    \multicolumn{2}{c|}{GCN}    & 1.63M   & 2.15M   & 3.42M & 1.85M   & 3.20M & 1.99M    \\ 
    \midrule
    \multirow{4}{*}{$p=2$} & SliceGCN & 953.87K & 1.21M   & 1.85M & 1.06M   & 1.74M & 1.14M    \\
                         & -SE       & 954.39K & 1.21M   & 1.85M & 1.06M   & 1.74M & 1.14M    \\
                         & -FF       & 1.09M   & 1.35M   & 1.85M & 1.06M   & 1.87M & 1.17M    \\
                         & -FFSE     & 1.09M   & 1.35M   & 1.85M & 1.06M   & 1.87M & 1.17M    \\ 
    \midrule
    \multirow{4}{*}{$p=3$} & SliceGCN & 730.00K & 899.80K & 1.33M & 798.74K & 1.25M & 858.25K  \\
                         & -SE       & 730.51K & 900.31K & 1.33M & 799.25K & 1.25M & 858.76K  \\
                         & -FF       & 850.70K & 1.02M   & 1.33M & 798.89K & 1.37M & 880.31K  \\
                         & -FFSE     & 851.21K & 1.02M   & 1.33M & 799.41K & 1.37M & 880.82K  \\
    \bottomrule
    \end{tabular}
    }
    \label{tab:params-stat-slice-gcn}
\end{table}

\begin{table}[!htbp]
    \caption{Experimental results of SliceGCN on different datasets.}
    \centering
    \resizebox{\linewidth}{!}
    {
    \begin{tabular}{c|lcc|lcc} 
    \toprule
    \multirow{2}{*}{Dataset}                        & Method          & Accuracy & Throughput   & Method          & Accuracy & Throughput    \\
                                                    &  $p=2$             &   & (epochs / s) &        $p=3$       &   & (epochs / s) \\  
    \midrule
    \multirow{5}{*}{Roman-Empire} & GCN     & 76.81   & \textbf{34.80}   &             &       &        \\ \cmidrule{2-7}
                                  & SliceGCN  & 77.30   & 25.23   & SliceGCN  & 77.21    & 20.55   \\
                                  & -SE       & 76.81   & \underline{26.93 }  & -SE       & 77.80    & 20.70   \\
                                  & -FF       & 79.74   & 23.17   & -FF       & 79.65    & 18.74   \\
                                  & -FFSE     & \underline{79.93}   & 22.71   & -FFSE     & \textbf{80.18}    & 18.57   \\
                                  \midrule
    \multirow{5}{*}{Amazon-Ratings} & GCN     & 48.37    & \textbf{26.10}    &             &       &        \\ \cmidrule{2-7}
                                    & SliceGCN  & 47.71    &  \underline{23.38}   & SliceGCN  & 46.95    & 17.54    \\
                                    & -SE       & 47.89    &  20.16   & -SE       & 46.1     & 16.95    \\
                                    & -FF       & \textbf{49.67}    &  18.77   & -FF       & \underline{48.77}    & 15.51    \\
                                    & -FFSE     & 47.59    &  18.44   & -FFSE     & 46.37    & 14.21    \\
                                    \midrule
    \multirow{5}{*}{Minesweeper} & GCN     &\underline{92.90}    &\textbf{25.90}    &             &       &        \\ \cmidrule{2-7}
                                & SliceGCN  & 92.47   & 18.12   & SliceGCN  & 92.53     & 12.62   \\
                                & -SE       & 92.66   & \underline{18.37}   & -SE       & 92.93     & 13.46   \\
                                & -FF       & 92.85   & 17.79   & -FF       & \textbf{93.12}     & 13.02   \\
                                & -FFSE     & 92.31   & 17.12   & -FFSE     & 90.96     & 13.03   \\
                                \midrule
    \multirow{5}{*}{Tolokers} & GCN     &82.94   &\textbf{31.24}    &             &       &        \\ \cmidrule{2-7}
                                    & SliceGCN  & 83.39  &\underline{ 25.81 }  & SliceGCN  &\underline{84.20}     &18.19   \\
                                    & -SE       & 84.06  & 25.36   & -SE       &\textbf{84.31}     &20.03   \\
                                    & -FF       & 83.34  & 25.24   & -FF       &82.78     &19.50   \\
                                    & -FFSE     & 82.57  & 25.24   & -FFSE     &83.08     &19.40   \\
                                    \midrule  
    \multirow{5}{*}{Questions} & GCN     &   76.61    &9.10    &             &       &        \\ \cmidrule{2-7}
                                & SliceGCN  & 76.62    &\textbf{10.87}    & SliceGCN  &77.92    &9.28   \\
                                & -SE       & \underline{77.94}    &\underline{10.12}    & -SE       &76.60    &9.01   \\
                                & -FF       & 77.35    &8.92    & -FF       & \textbf{78.05}    &7.73   \\
                                & -FFSE     & 75.50    &8.86    & -FFSE     & 77.24    &7.52   \\
                                \midrule
    \multirow{5}{*}{OGBN-Arxiv} & GCN     & \textbf{69.74}     &4.15    &             &       &        \\ \cmidrule{2-7}
                                    & SliceGCN  &69.25     &2.81    & SliceGCN  & \underline{69.59}     &\textbf{4.37}   \\
                                    & -SE       &68.82     &2.71    & -SE       & 68.85     &\underline{4.33}   \\
                                    & -FF       &68.00     &2.62    & -FF       & 67.78     &3.86   \\
                                    & -FFSE     &67.10     &2.59    & -FFSE     & 66.68     &3.82   \\
    \bottomrule
    \end{tabular}
    }
    \label{tab:res-slice-gcn}
\end{table}

Overall, in terms of accuracy, SliceGCN and its variants not only maintain accuracy compared to GCN on a single GPU, but also achieve better convergence—resulting in various degrees of accuracy improvement on the test set. In terms of throughput, experiments show that SliceGCN is more efficient on large-scale datasets—except for the Questions dataset, the throughput of SliceGCN and its variants decreased across the five larger heterophilic graph datasets. However, it is worth noting that the Questions dataset has the largest number of nodes, 2 to 5 times more than the others. The throughput of SliceGCN improved on the Questions dataset, and it also improved on a large-scale dataset, OGBN-Arxiv, where the node count is 3.5 times that of Questions and the number of edges is 10 times greater. Additionally, it can be observed that the throughput of GCN on a single GPU is similar on the Amazon-Ratings and Minesweeper datasets, but the decrease in throughput of SliceGCN on Amazon-Ratings is significantly smaller than on Minesweeper, and the node and edge counts of Amazon-Ratings are larger than those of Minesweeper, which aligns with the conclusion that SliceGCN performs better in terms of efficiency when the graph data scale is larger.

To ensure that slicing features does not affect model convergence, two designs were explored: feature fusion and slice encoding. In the experiments, except for the Tolokers and OGBN-Arxiv datasets, adding feature fusion generally improved the accuracy of SliceGCN, which indicates that splitting features and using multiple GCNs for message passing before concatenating the node representations indeed may cause a drop in accuracy. However, except on the OGBN-Arxiv dataset, SliceGCN itself can achieve accuracy comparable to that of single-GPU GCN, and after using feature fusion or slice encoding, the accuracy of SliceGCN can even be significantly higher. According to the statistics in Table \ref{tab:params-stat-slice-gcn}, SliceGCN and its variants have fewer parameters compared to single-GPU GCN in our experiments, suggesting that SliceGCN might constitute a parameter-efficient architecture similar to Mixture of Experts (MoE) \cite{moe}. However, we still need to validate it, to be equal, by conducting experiments on single GPU with less GCN parameters.

Furthermore, as illustrated in Figure \ref{fig:SliceGCN-p3} and Figure \ref{fig:SliceGCN-p2}, this paper visualizes the validation accuracy during training for GCN and SliceGCN and their variants on various datasets at $p=2$ and $p=3$. As can be seen, compared to the test set accuracy results taken at the highest validation set accuracy in Table \ref{tab:res-slice-gcn}, Figure \ref{fig:SliceGCN-p3} more intuitively shows the significant differences between single-GPU GCN and SliceGCN and its variants. Although higher accuracy on the validation set does not necessarily imply better performance in actual inference, it still reflects the differences in convergence capabilities of SliceGCN and its variants during training. Overall, single-GPU GCN and SliceGCN have the fastest convergence rates early in training, which visually appears to be because these two do not have additional MLPs for feature fusion or slice encoding to train; GCN converges faster on the Roman-Empire, Amazon-Ratings, and Questions datasets, while SliceGCN converges faster on the other three datasets. Interestingly, although the accuracy improvements from feature fusion and slice encoding are sometimes not significant in Table \ref{tab:res-slice-gcn}, Figure \ref{fig:SliceGCN-p3} shows that both feature fusion and slice encoding, whether used separately or together, can stabilize the model to varying degrees in the later stages of training, whereas single-GPU GCN is more prone to significant performance decline in the later training stages and more fluctuations throughout the training process. Although this can be interpreted as feature fusion or slice encoding introducing more modules making the model more complex and preventing overfitting, it is evident that the simpler-parameter SliceGCN can also maintain stability in the later stages of training without these two designs.

\section{Conclusion, Limitations, and Future Research}

This paper begins by outlining the research background of distributed GCNs and the motivation behind the proposed SliceGCN—to enable single computing devices to store as many local nodes as possible to minimize inter-device communication during distributed training. SliceGCN slices the input node features into multiple parts, allowing each computing node to handle smaller node features but, as a result, accommodate larger-scale graph data in the form of a complete graph structure. Experiments show that SliceGCN increases throughput on large-scale graph data and that such an architecture may possess parameter-efficient characteristics, achieving higher accuracy with a smaller parameter size. Additionally, for SliceGCN, this paper proposes and studies feature fusion and slice encoding to ensure convergence, exploring their roles in the experiments. It was found that these designs enable SliceGCN to achieve higher accuracy and make the model more stable in the later stages of training. These designs could also be explored in a single-machine context to enhance the performance of GNNs and even other deep learning models, warranting further investigation.

This paper is currently a preliminary exploration, and the feature-sliced distributed training proposed is evidently general, not limited to GNNs, although our initial intent was to explore distributed training for GNNs. Therefore, the method presented in this paper is still in its nascent form. In the future, we plan to continue our exploration in two directions: 1. Integrate it more organically with GNNs and conduct experiments on a broader range of GNN methods; 2. Study the slice-concat framework from a theoretical perspective to determine its applicability in generic deep learning scenarios beyond GNNs, and whether it can enhance training precision, convergence, and efficiency.

\bibliographystyle{unsrt}
\bibliography{ref.bib}

\end{document}